\documentclass[runningheads]{llncs}
\usepackage[T1]{fontenc}
\usepackage{graphicx}
\usepackage{fontspec}
\usepackage{enumitem}
\usepackage{multirow}
\usepackage{cite}
\usepackage{orcidlink}

\usepackage{array}
\usepackage{booktabs} 
\usepackage{hyperref}
\hypersetup{
    colorlinks=false,   
    pdfborder={0 0 0}   
}

\usepackage{fontspec}
\usepackage{polyglossia}
\newfontfamily\bengalifont[
  Script=Bengali,
  Path= ./ ,
  Extension=.ttf
]{Kalpurush}
\usepackage{fontspec}
\usepackage{polyglossia}
\setdefaultlanguage{english}
\setotherlanguage{bengali}
\usepackage{cite}
\usepackage{amsmath,amssymb,amsfonts}
\usepackage{array}
\usepackage{graphicx}
\usepackage{textcomp}
\usepackage{xcolor}
\usepackage{amsmath}
\usepackage{tikz}
\usepackage{nth}
\usepackage{subcaption} 
\usepackage{fancyhdr}
\usepackage{float}
\usepackage{csquotes}

\usepackage{comment}
\usepackage{algorithmicx,algpseudocode}
\usepackage{polyglossia}
\usepackage[linesnumbered, ruled, vlined]{algorithm2e}

\usepackage{siunitx,array,multirow}

\usepackage{tabularx}
\usepackage{multirow}
\usepackage{threeparttable}
\usepackage{multicol}
\usepackage{xcolor,colortbl}
\usepackage{amsmath}

\usetikzlibrary{shapes,arrows}
\usepackage{verbatim}

\usepackage{url}
\usepackage{amsmath}
\usepackage{textcomp}
\usepackage{siunitx}
\usepackage[utf8]{inputenc}
\usepackage{upgreek}

\makeatletter
\def\ps@firstpagefooter{%
  \def\@oddhead{}%
  \def\@evenhead{}%
  \def\@oddfoot{%
  \parbox[t]{0.9\textwidth}{\small
  Accepted at 3rd International Conference on Big Data, IoT and Machine Learning (BIM 2025)\\
  September 25--27, 2025
  }\hfill\thepage}%
  \def\@evenfoot{\@oddfoot}}
\makeatother
 
\begin{document}
\title{BanglaMed-QA: A Question Answering System for Healthcare Support in Bangla}

\titlerunning{BanglaMed-QA}

%

\author{Rowzatul Zannat\inst{1}\orcidlink{0009-0004-9261-1040} 
\and Abdullah Al Shafi\inst{2} \orcidlink{0009-0004-3082-9749}
\and K. M. Azharul Hasan\inst{3}\orcidlink{0000-0003-1228-9043} 
\and Atia Shahnaz Ipa\inst{4} \orcidlink{0009-0006-7308-899X}
}
\authorrunning{R. Zannat et al.}

\institute{Khulna University of Engineering \& Technology, Khulna-9203, Bangladesh
\email{w.rzrowza@gmail.com}\inst{1}, \email{abdullah@iict.kuet.ac.bd}\inst{2},  \email{az@cse.kuet.ac.bd}\inst{3} \email{atia.s.ipa@gmail.com}\inst{4}}

\maketitle
\thispagestyle{firstpagefooter}

\begin{abstract}
Medical question answering (QA) systems have become crucial tools for providing reliable health information. But they remain very unexplored for low-resource languages like Bangla due to limited datasets and systems tailored to these languages. To address this, we introduce BanglaMed-QA, a robust QA system specifically designed for the Bangla medical domain. The process begins with building a structured medical knowledge base that includes 4,493 QA pairs in 9 categories under 506 diseases. To improve semantic comprehension, domain-specific root word dictionaries and synonym sets are proposed, in addition to part-of-speech tagging for anaphora resolution. We adopt supervised machine learning models in which SVM is found to be the best model to categorize questions. Multiple similarity metrics, including cosine, Jaccard, BM25, and Levenshtein, are applied with soft and hard voting methods for query matching. The performance of the QA system has been evaluated in two aspects, with a 95\% F1 score in an automated evaluation and an average human satisfaction rating of 0.9 out of 1.0. This validates the real-world application of BanglaMed-QA in closing the healthcare information gap for Bangla speakers.

\end{abstract}

\section{Introduction}
\label{sec:introduction}
One of the most valuable aspects of life is health, and it is natural for people to seek answers when they feel ill or are uncertain about their condition \cite{sen2024healthcare}. However, access to reliable and understandable medical information remains beyond the reach of most people, particularly those in areas where English is not the most spoken language \cite{rahman2019disha}. The question-answering (QA) systems proved to be of great use in this context. Instant responses with accurate and correct solutions to a particular question are the ones for which the user would normally prefer \cite{singhal2025toward}. Instead of searching for a long document, people want to receive only correct, precise, and summarized information. The problem is magnified when the question is in low-resource languages such as Bangla, and sometimes the desired information might not be found even after scanning the whole document \cite{sen2024healthcare}. Thus, a QA system for the Bangla language in the healthcare field can be extremely beneficial in addressing these issues. But such works are very limited \cite{sen2024healthcare}. In this work, our goal is to mitigate the gap and build a robust and user-friendly quality assurance system for healthcare support. Our contributions can be summarized as follows.
\begin{itemize}
\item We have constructed a structured knowledge base of the medical domain in Bangla to develop a healthcare support system, as there is a scarcity of such a corpus in the Bangla medical literature.

\item Using POS tagging, we have developed an anaphora resolution module that successfully resolves pronouns to their corresponding nouns, suitable for a medical QA system.

\item We have explored various supervised machine learning models as our question classifier to speed up the similarity matching task.

\item We have used cosine similarity \cite{rahman2019disha}, Jaccard similarity \cite{abd2025enhancing}, BM25 \cite{bronda2024integrating}, and Levenshtein similarity \cite{zhang2025cecld}, and then developed a soft and hard voting ensemble of these measures of similarity to find the similarity between the user's question and knowledge base and output the answer of the most similar question with a threshold.

\item We have also developed a practical system with an interactive user interface (UI) to visualize the usefulness of our proposed system.
\end{itemize}

The paper is organized as follows, Sections \ref{sec:introduction} provides the importance of the QA system in the medical domain. Section \ref{sec:literature} compares some existing works. Section \ref{sec:dataset} presents a Bangla Question Answering dataset in the medical area. The suggested method to create the \textquote*{BanglaMed-QA} system is described in Section \ref{sec:methodology}. The experimental analysis and results are presented in Section \ref{sec:result}. Finally, a conclusion is drawn in Section \ref{sec:conclusion}.

\section{Literature Review}
\label{sec:literature}
In this section, we review the landscape of research on the medical question-answering system in the context of methods, datasets, and evaluation approaches that contribute to the field. First, we discuss rule-based, UML, information retrieval(IR), and knowledge graph(KG) based models. We then move on to the recent progress made possible with neural networks and hybrid methods.

Jacquemart et al. \cite{jacquemart2003towards} conducted an early feasibility study in medical QA systems to model questions as UMLS-based triples. Although 90\% of the student questions followed this structure, there appeared to be issues related to the normalization and scalability of the questions. Abacha et al.\cite{abacha2015means} created MEANS, a system that mixes NLP and semantic web tools, SPARQL, and query relaxation, although it suffered from scalability. Feng et al. \cite{feng2018chinese}  presented CQASMD, a Chinese QA system that took advantage of large pairs of QA, FastText-based classification and similarities in disease-symptom vectors. Huang et al. \cite{huang2021knowledge} used weighted path classification in conjunction with knowledge graph construction.

In the Bangla context, Rahman et al. \cite{rahman2019disha} developed a Bangla chatbot for healthcare care, using supervised learning to predict diseases. For query matching, TF-IDF with cosine similarity was used and an English data set was selected to translate it into Bangla, thereby addressing the issue of data scarcity. Sen et al. \cite{sen2024healthcare} proposed a Bengali medical QA system for the interpretation of prescriptions and reports; however, it suffers due to the size of the data and the lack of clinical testing. \cite{rahman2021intelligent} developed a Bangla chatbot, namely TUNI, that interacts with the audience for psychological issues with simple sentences. The authors in \cite{zannat2025bridging} proposed a medical helper, which was trained on a custom dataset \cite{ratul2025structured}, can predict disease based on given symptoms. Recently, Maharjan et al.\cite{maharjan2024openmedlm} proposed OpenMedLM, a framework that helps open source LLMs achieve stronger performance in medical quality assurance benchmarks by few shots, self-consistency, etc., but remains limited to multiple choice settings and lacks validation under clinical conditions.

Despite these developments, issues such as limited annotated datasets, sophisticated work in Bangla, and the need for real-world validation persist.
\section{A Bangla Question Answering Dataset for Healthcare Support}
\label{sec:dataset}
The workflow started with a comprehensive study and a data collection phase. We waded through dozens of online data sources, from health blogs to Wikipedia to newspapers, and verified social media pages and groups. We diligently assembled the raw data collection and then cleaned it based on quality, relevance, and ultimately meaning.

\begin{figure}[htbp]
\centering
     \begin{subfigure}[b]{0.43\textwidth}
         \centering
         \includegraphics[width=\textwidth]{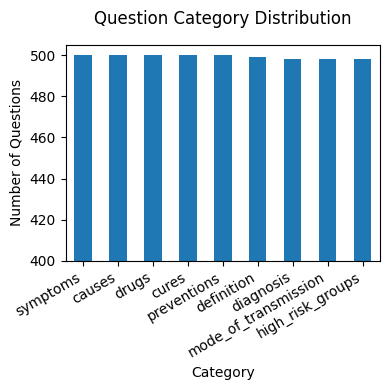}
         \caption{}
         \label{question_cat}
     \end{subfigure}
     \hspace{0.05\textwidth}
     \begin{subfigure}[b]{0.44\textwidth}
         \centering
         \includegraphics[width=\textwidth]{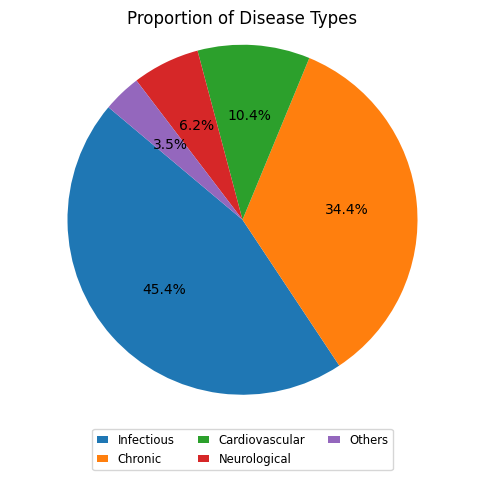}
         \caption{}
         \label{disease type}
     \end{subfigure}
        \caption{(a) Distribution of question categories in our dataset. (b) The proportion of types of diseases covered in the study.
}
\label{qx+dt}
\end{figure}

\textbf{Question Answer Pair Dataset with Category: }The dataset consists of three columns, namely \textquote*{Question}, \textquote*{Answer}, and \textquote*{Category}. The \textquote*{Question} column represents a query related to healthcare, and the \textquote*{Answer} column represents the answer to the corresponding query. There are a total of 4493 QA pairs with 9 categories related to 506 diseases. Table \ref{tab:QA_pair} shows a sample portion of our medical QA pair dataset. The distribution in Fig. \ref{question_cat} shows that all categories have an almost equal number of questions, thus a balanced dataset. Fig. \ref{disease type} shows that maximum diseases are infectious and chronic.

\begin{table*}[!ht]
\caption{A sample portion of our Question Answer Pair dataset in the Medical Domain. Here, the samples are about rabies (\textbengali{জলাতঙ্ক}).} 
\centering 
\begin{tabular}{|p{2.4cm}|p{8.6cm}|p{1.5cm}|}
\hline
Question & Answer & Category\\
\hline 
\textbengali{জলাতঙ্ক কি?} (What is rabies?) & \textbengali{জলাতঙ্ক একটি তীব্র ভাইরাল সংক্রমণ। এটি খামার বা বন্য প্রাণীদের দ্বারা সংক্রামিত হয়; সাধারণত মাংসাশী যেমন কুকুর, বিড়াল, শেয়াল, রেকুন।(Rabies is an acute viral infection. It is spread by farm or wild animals, typically carnivores including foxes, raccoons, dogs, and cats.)} & definition\\
\hline
\textbengali{জলাতঙ্ক কিভাবে প্রতিরোধ করা যায়?} (How can rabies be prevented?) & \textbengali{জলাতঙ্ক প্রতিরোধে পোষা ও অ-পোষা সব বিড়াল-কুকুরকে জলাতঙ্কের টিকা কার্যক্রমের আওতায় আনাও একটি কার্যকর উপায়।} (Bringing all domestic and non-domestic cats and dogs under the rabies vaccination program is also an effective way to prevent rabies.) & preventions\\
\hline
\end{tabular} 
\label{tab:QA_pair} 
\end{table*}

\begin{table}[h!]
\centering
\caption{A part of the list of Synonym sets. Any synonyms will be replaced by the selected word, which helps further processing.}
\begin{tabular}{|p{7cm}|l|p{1.6cm}|}
\hline
\textbf{Synonyms} & \textbf{Meaning} & \textbf{Replaced By}\\ \hline
{\textbengali{লক্ষণ, চিহ্ন, পরিচয়, আভাস, নিদর্শন, উপসর্গ, সনাক্ত}} & Symptoms & \textbengali{লক্ষণ}\\ \hline
{\textbengali{প্রতিকার, আরোগ্য, নিবারণ, প্রশমিতকরণ, দূরীকরণ ,চিকিৎসা}} & Cure & \textbengali{প্রতিকার}\\ \hline
{\textbengali{ডায়েবেটিস, বহুমুত্র, ডায়েবেটিক, ডায়বেটিস, ডায়বেটিক}} & Diabetic & \textbengali{ডায়েবেটিস}\\ \hline
\end{tabular}
\label{tab:syn}
\end{table}

\textbf{POS tagging dataset: }We have done the POS tagging of the questions of the QA pair dataset for anaphora resolution. The POS tagging is done like this: \textbengali{ডেঙ্গু/NN-DIS কি/PRP ?/Sym} (What/PRP is Dengue/NN-DIS ?/Sym).

\textbf{Bangla root word dictionary in medical domain: }There is no standard root word dictionary of the medical domain in Bangla. So, we have developed such a dictionary. The dictionary has 617 words. Some examples are \textbengali{ফুসকুড়ি, ক্ষুধা, ডেঙ্গু, রোগ, কান}, etc. 

\textbf{Synonym sets: }We have made a list of synonym sets where each set represents synonym words to handle variations of synonyms to ask the same query. Table \ref{tab:syn} shows a portion of the root word list.

\section{A QA System for Medical Bangla Text}
\label{sec:methodology}
\begin{figure*}[ht]
\centering
\centerline{\includegraphics[scale=.5]{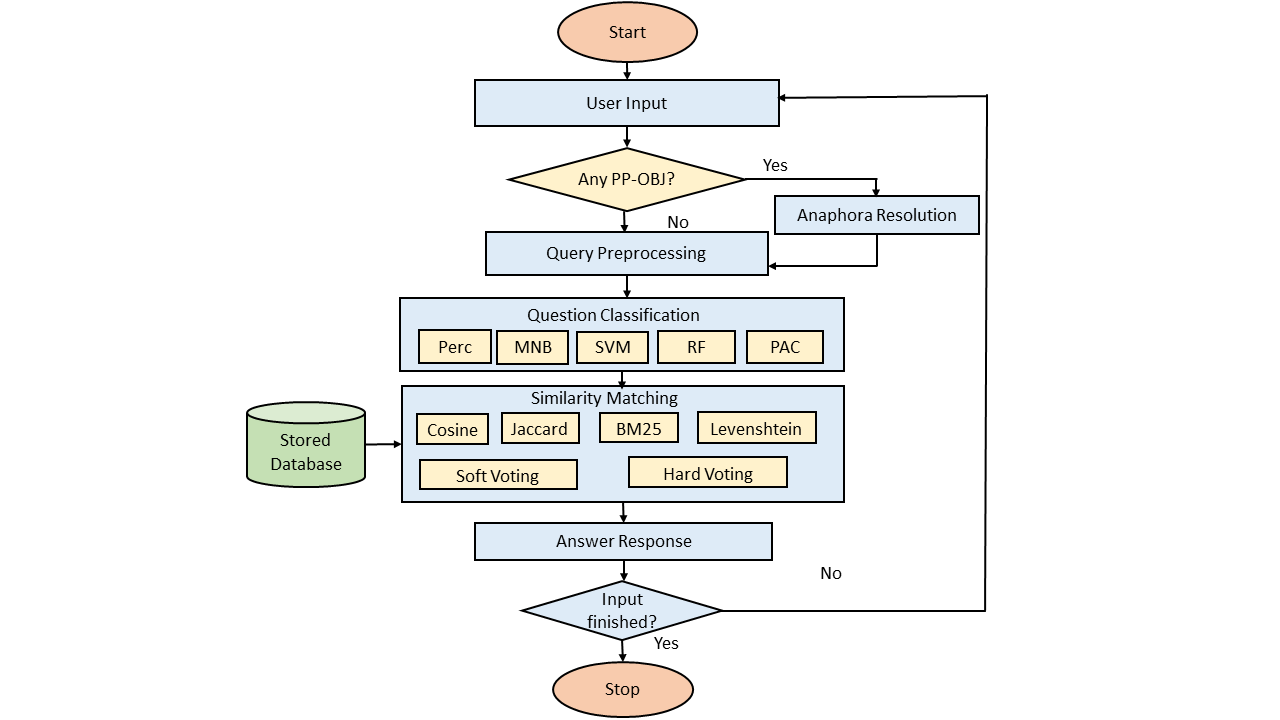}} 
\caption{Flowchart of our proposed \textquote*{BanglaMed-QA} system}
\label{fig:proposed}
\end{figure*}

Fig. \ref{fig:proposed} shows the workflow of our proposed \textquote*{BanglaMed-QA} system. It begins with user input, checks for pronoun-based objects (PP-OBJ) to resolve anaphora, preprocesses and classifies the query, then matches it with a database to generate an answer. The loop continues until the input is finished.

\subsection{Data Preprocessing}
\begin{figure*}[ht]
\centering
\centerline{\includegraphics[width=0.94\textwidth]{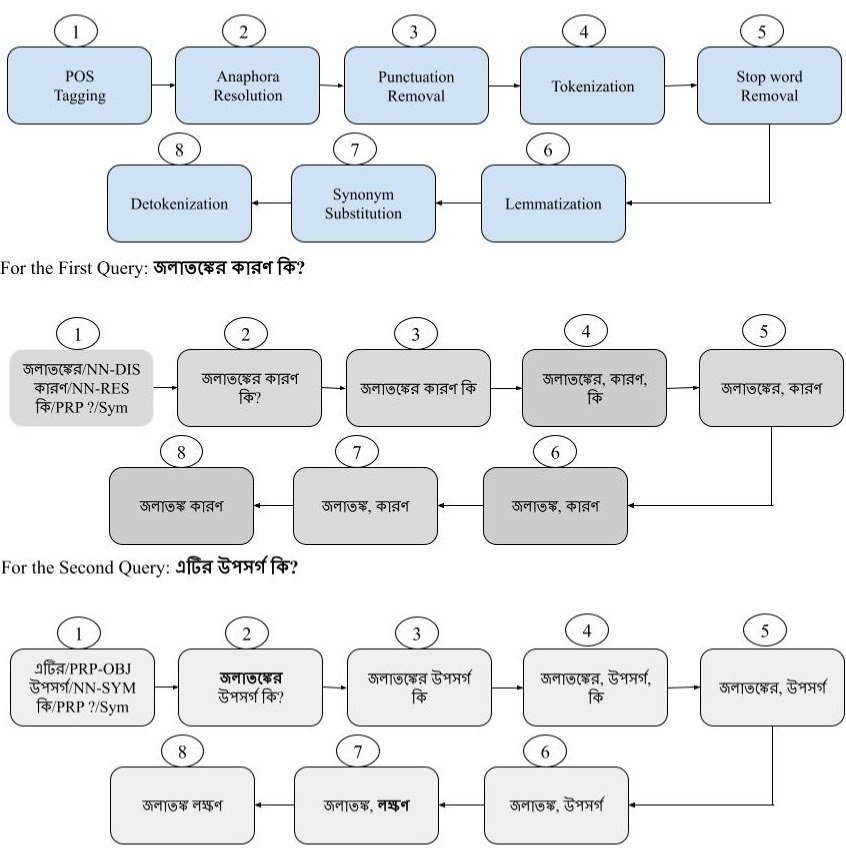}}
\caption{Query Preproceesing Steps.}
\label{fig:preprocessing}
\end{figure*}

BanglaMed-QA applies several preprocessing methods to effectively prepare the data. First, we have performed POS tagging using the TnT tagger\cite{brants2000tnt}, basic anaphora resolution by tracking disease nouns, and punctuation removal. Then, we use tokenization, stop word removal, and custom lemmatization for key term extraction, followed by synonym substitution for better semantic matching and detokenization to formulate the final queries. Fig. \ref{fig:preprocessing} shows the steps of pre-processing with examples.


\subsection{Question Categorization}
After preprocessing, we have categorized the input question into one of the 9 categories so that the question matching methods become fast as the number of question pairs to check similarity would be reduced. From our QA pair training dataset, We take \textquote*{Question}  as a feature and \textquote*{Category} as a class. At first, We use a TF-IDF vectorizer with an n-gram range of (1, 2) to convert text into numerical feature representations. This setup allows the model to capture unigrams, and bigrams, enabling it to consider both individual words and short phrases, which is extremely beneficial in our case. Although not implementing explicit backoff, this configuration implicitly provides a form of backoff by including lower-order n-grams when higher-order n-grams are sparse or absent. After that, we have explored 5(five) different supervised machine learning methods, including Perceptron (Perc)\cite{zannat2025bridging}, Multinomial Naive Bayes(MNB)\cite{ahmed2025blockchain}, Passive Aggressive Classifier(PAC)\cite{rahman2019disha}, Random Forest(RF)\cite{imani2025comprehensive}, and Support Vector Machine(SVM)\cite{yang2025deep}. For each of the methods except MNB, we have used hyperparameter tuning with 5-fold cross-validation. All the methods perform well (Section \ref{sec:result}) and we finally choose SVM as our question classifier.

\subsection{Question Matching Methods}
For the explanation of question matching algorithms, we take \textbengali{স্ট্রোক প্রতিকার} (stroke cure) as an example processed query (Q) as well as one of the stored queries ($S_1$). And $S_i$ denotes the i-th stored query. All the values given below are taken by running the program (actual value).

\subsubsection{Base Similarity Methods\\}
Four base similarity methods have been explored in the experiment.

\textbf{TF-IDF with cosine similarity: }This method \cite{rahman2019disha} represents each question using TF-IDF with ngram range (1,2) to emphasize important unigrams and bigrams, and then measures cosine similarity (C) between their vector representations.
\begin{equation}
\label{c}
C(Q,S_i) = \cos(\theta) = \frac{\vec{Q} \cdot \vec{S_i}}{\|\vec{Q}\| * \|\vec{S_i}\|}
\end{equation}
For our example, \\
\text{IDF \textbengali{(স্ট্রোক} (stroke))} = 1.693, \text{IDF \textbengali{(প্রতিকার} (cure))} = 1.693, \text{IDF \textbengali{(স্ট্রোক প্রতিকার} (stroke cure))} = 1.945\\
TF-IDF (Q) = [0,...., 0.323474, 0, ....,599035,0,....,0.732476], and TF-IDF ($S_1$) = [0,...., 0.323474, 0, ....,599035,0,....,0.732476].
\\So using Eq. \ref{c},
\[
C (Q, S_1) = \frac{1.000005}{\sqrt{1.000005} \cdot \sqrt{1.000005}} = 1.0
\]

\textbf{Jaccard similarity: }Jaccard similarity(J) \cite{abd2025enhancing} measures lexical overlap by dividing the number of words shared between two questions by the total unique words. 
\begin{equation}
\label{j}
J(Q,S_i) = \frac{|Q \cap S_i|}{|Q \cup S_i|}
\end{equation}

\begin{center}
\begin{tikzpicture}
  \begin{scope}
    \fill[blue!30, opacity=0.5] (0,0) circle (1.2);
    \fill[red!30, opacity=0.5] (1.1,0) circle (1.2); 
  \end{scope}

  \node at (0,1.5) {\text{Query}};
  \node at (1.4,1.5) {\text{Stored}};

  \node at (-0.5,0.2) {\scriptsize };
  \node at (1.9,0.2) {\scriptsize };
  \node at (0.55,0) {\scriptsize \textbengali{স্ট্রোক প্রতিকার}}; 

  \node at (0.7,-1.8) {
   Using Eq. \ref{j}, $J(Q, S_1) = \frac{|Q \cap S_1|}{|Q \cup S_1|} = \frac{2}{2} = 1.0$
  };
\end{tikzpicture}
\end{center}

\textbf{BM25 similarity: }BM25 \cite{bronda2024integrating} is a probabilistic retrieval model that ranks documents using term frequency, inverse document frequency, and document length. Let, \(B(Q,S_i)\) be the raw BM25 score between two questions and \(B_{\text{max}} \) be the maximum BM25 score observed across all question pairs and $B_{\text{norm}}(Q,S_i)$ be the normalized BM25 similarity.

\begin{equation}
\label{b}
\text{BM25}(Q, S_i) = \sum_{j=1}^{n} \text{IDF}(q_j) \cdot \frac{f(q_j, S_i) \cdot (k + 1)}{f(q_j, S_i) + k \cdot \left(1 - b + b \cdot \frac{|S_i|}{\text{avgdl}} \right)}
\end{equation}
\begin{equation}
\label{bn}
B_{\text{norm}}(Q,S_i) = \frac{B(Q,S_i)}{B_{\text{max}}} \quad \text{if } B_{\text{max}} > 0, \text{ else } 0   
\end{equation}

Here, $q_j$ is each term of Q. And we set k = 1.5 and b=0.75 in our experiment.
For our example, term contributions are : 
\\\textbengali{স্ট্রোক (stroke)}: $1.693 \times 1.1765 = 1.992$, \textbengali{প্রতিকার (cure)}: $1.693 \times 1.1765 = 1.992$, \textbengali{স্ট্রোক প্রতিকার (stroke cure)}: $1.945 \times 1.1765 = 2.288$ 

So, using Eq. \ref{b} and \ref{bn}, $BM25(Q,S_1)$ = 1.992 + 1.992 + 2.288 = 6.272 and \\
$B_{norm}(Q,S_1)$ = 1.0

\textbf{Levenshtein similarity: }
This method calculates similarity by counting the minimum edits (insertions, deletions, substitutions) needed to transform one string into another, normalized between 0 and 1 \cite{abd2025enhancing}. Let \( L(Q,S_i) \), \( L_d(Q,S_i) \), and \( M \) denote Levenshtein similarity, Levenshtein distance, and maximum length between questions Q and $S_i$ respectively, where \( M = \max(|Q|, |S_i|) \).
\begin{equation}
L(Q,S_i) = 1 - \frac{L_d(Q,S_i)}{M} \quad \text{if } M > 0, \text{ else } 0    
\end{equation}

\[
\text{$L_d(Q,S_1)$ = 0} \Rightarrow L(Q,S_1) = 1 - \frac{0}{17} = 1.0; \text{ Here, M = 17}
\]

\subsubsection{Ensemble Similarity Methods\\}
\textbf{Soft voting similarity: }For the final scoring (S) of the similarity between two sentences, the average of the four similarity measures is taken to combine their strengths for a more balanced result. In the case where the final score is greater than an appropriate threshold, the two sentences are considered similar.
\begin{equation}
S(Q,S_i) = \frac{C(Q,S_i) + J(Q,S_i) + L(Q,S_i) + B_{\text{norm}}(Q,S_i)}{4}
\end{equation}

\[
\text{Thus, } S(Q,S_1)= \frac{1.0 + 1.0 + 1.0 + 1.0}{4} = \frac{4}{4} = 1.0
\] \\

If \( S \geq \tau \), then the sentences are considered similar, where \( \tau \) is the predefined similarity threshold. We set \( \tau = 0.8 \).\\

\textbf{Hard voting similarity: }Under the hard voting (H) strategy, individual similarity measures cast their votes (binary) depending on whether they exceed a set threshold. If the majority votes, meaning 3 or even all 4 votes, deeming the pair as similar, then indeed it is considered similar.

Let \( sim_i \) be the normalized similarity scores for the methods: cosine, Jaccard, Levenshtein, and BM25. The voting logic is defined as follows:
\begin{equation}
m_i =
\begin{cases}
1, & \text{if } sim_i \geq T \\
0, & \text{otherwise}
\end{cases}
\end{equation}

\begin{equation}
H(Q,S_i) =
\begin{cases}
1, & \text{if } \sum_i m_i \geq 3 \\
0, & \text{otherwise}
\end{cases}
\end{equation}
Here, \( T \) denotes the threshold value for decision making. We set \( T \) = 0.8.\\

Since all four models produce the same prediction ($m_i = 1$ for all $i$), the cumulative vote count is 4 out of 4, resulting in a unanimous hard vote decision of 1.

\section{Experimental setup, Results and Analysis}
\label{sec:result}
\subsection{Evaluation Strategy}
To evaluate our question categorization methods, we divide our \textquote*{QA pair with category} dataset into two parts: 80\% for training and 20\% for testing. Since there was no publicly available dataset aligned with the requirements of our overall system, we have manually constructed a test set to evaluate the performance and also performed real-time testing through human interaction. The manually crafted data set consists of 880 input queries paired with their corresponding target responses and categories from the stored collection. Special attention was paid to include challenging cases that involve synonym variations, phrased expressions, and domain-specific terminology to analyze the effectiveness of the system under realistic usage. To evaluate the applicability of our Bangla Medical QA system in the real world, we performed a human evaluation using an interactive form. Table \ref{table:human_eval} illustrates sample evaluation records in which human participants provided questions, reviewed system responses, and rated between 0 and 1 according to their expectations. A group of 10 evaluators (students) manually entered Bangla medical-related questions. Each evaluator asked 10 questions. These questions reflect natural and practical queries that users might ask in real-world scenarios.

\subsection{Evaluation Metrics}

Important factors to consider when evaluating question categorization models include accuracy, precision, recall, and F1-score. For evaluation of our overall system on the manually crafted dataset, we have treated each transformed query of the test set as a separate feature vector, and predictions that do not match any of them are classified as a separate class, NO\_PREDICTION. Coverage is defined as how many predictions are non-None by the model for a given number of examples, indicating how often the model answers to answer. Accuracy, precision, and recall are calculated as the number of examples for which the predicted label is equal to the true label out of the total and NO\_PREDICTION is considered to be an incorrect case. F1-score is calculated using the harmonic mean of precision and recall.

\begin{table*}[!ht]
\caption{Sample Human Evaluation Record. Each person asked a question using our \textquote*{BanglaMed-QA} and the system answers. The peer then rates from 0 to 1 based on his/her satisfaction.}
\centering
\begin{tabular}{|p{1cm}|p{2.3cm}|p{6.7cm}|c|}
\hline
\textbf{Person} & \textbf{Question} & \textbf{System Answer} & \textbf{Rating(0-1)} \\
\hline
P1 & \textbengali{ডায়াবেটিসের উপসর্গ কী কী?(What are the symptoms of diabetes?)} & \textbengali{ডায়েবেটিস এর লক্ষণ অধিক ক্ষুধা ,অতিরিক্ত মুত্রনিষ্কাসন, চোখে সমস্যা,অস্থিবাধা এবং জড়তা। (Symptoms of diabetes include increased hunger, excessive urination, eye problems, bone loss, and numbness.)}  & 0.95 \\
\hline
P2 & \textbengali{বহুমূত্র কি? (What is diabetic?)} & \textbengali{ডায়েবেটিস একটি ক্রনিক মেডিক্যাল কন্ডিশন বা রোগ,যেখানে রক্তের গ্লুকোজ স্তর অস্বাভাবিক ভাবে বাড়তে থাকে। (Diabetic is a chronic medical condition or disease in which blood glucose levels increase abnormally.)} & 0.9 \\
\hline
\end{tabular}
\label{table:human_eval}
\end{table*}

\subsection{Quantitative Results}

Table \ref{tab:models_parameters} shows the optimal hyperparameters of the models for question categorization. The values were obtained by first performing baseline validation to find initial values and then using 5-fold cross-validation with grid search.

\begin{table}
\centering
\caption{Result of Hyperparameter Tuning with 5-Fold Cross-Validation. Here, C is the regularization parameter.}
\begin{tabular}{|l|l|l|}
\hline
\textbf{Methods} & \textbf{Hyperparameters} & \textbf{Optimal Value} \\ \hline
Perc & alpha & 0.0001 \\ \cline{2-3} & penalty & l1 \\ \hline
PAC & C & 1 \\ \cline{2-3} & loss & squared hinge \\ \cline{2-3} & max\_iter & 500\\ \hline
RF & n\_estimators & 40 \\ \cline{2-3} & max depth & 50 \\ \cline{2-3} & min samples leaf & 2 \\ \hline
SVM & kernel & linear \\ \cline{2-3} & C & 1 \\ \hline
\end{tabular}
\label{tab:models_parameters}
\end{table}

\begin{figure}
\centering
     \begin{subfigure}[b]{0.5\textwidth}
         \centering
         \includegraphics[width=\textwidth]{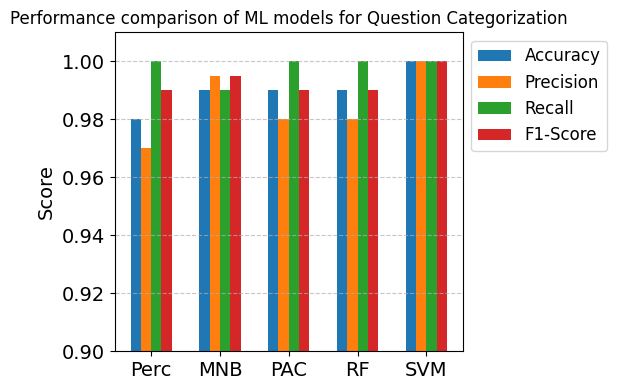}
         \caption{}
         \label{performance_comp}
     \end{subfigure}
     \hspace{0.05\textwidth}
     \begin{subfigure}[b]{0.42\textwidth}
         \centering
         \includegraphics[width=\textwidth]{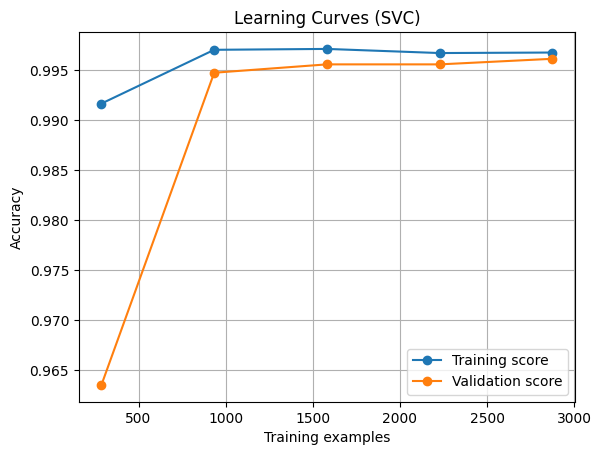}
         \caption{}
         \label{svm_lc}
     \end{subfigure}
        \caption{(a) Performance comparison of various supervised machine learning methods for question categorization. (b) Learning curve of SVM.}
\label{qx+dt}
\end{figure}

Fig. \ref{performance_comp} shows that all models performed exceptionally well in the categorization of the questions, with scores near 1.0 on all metrics. SVM showed the best overall performance, closely followed by RF and PAC. Minimal performance variation indicates high reliability between the selected models. The convergence pattern in Fig. \ref{svm_lc} indicates a well-fitted model that is neither over-fitted nor under-fitted. This also applies to other models that we employed in our research.

From the experiment, we have also observed that this categorization speeds up the question similarity matching approach by around one-ninth (1/9).

\begin{table}
\centering
\caption{Performance of our proposed method on the manually crafted test dataset.}
\begin{tabular}{|c|c|c|c|c|c|}
\hline
\textbf{Similarity method} & \textbf{coverage} & \textbf{Accuracy} & \textbf{Precision} & \textbf{Recall} & \textbf{F1-score}\\ \hline
Cosine Similarity & 1.0 & 0.94 & 0.94 & 0.94 & 0.94\\ \hline
Jaccard Similarity & 1.0 & 0.94 & 0.94 & 0.94 & 0.94\\ \hline
BM25 Similarity & 1.0 & 0.94 & 0.92 & 0.94 & 0.93\\ \hline
Levenshtein Similarity & 1.0 & 0.94 & 0.94 & 0.94 & 0.94\\ \hline
\textbf{Soft Voting Similarity} & 1.0 & \textbf{0.95} & \textbf{0.95} & \textbf{0.95} & \textbf{0.95}\\ \hline
Hard Voting Similarity & 0.94 & 0.94 & 0.94 & 0.94 & 0.94\\ \hline
\end{tabular}
\label{tab:test_result}
\end{table}

\begin{figure}
\centering
\centerline{\includegraphics[width=0.52\textwidth]{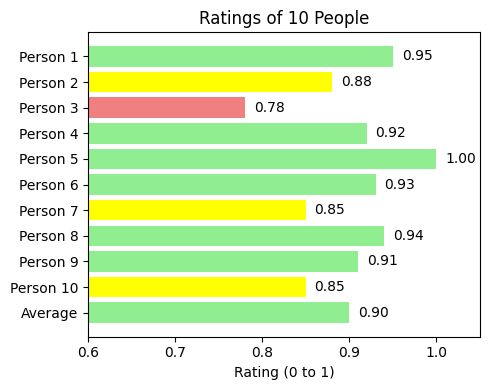}}
\caption{Human evaluation of 10 individuals rating (0-1)along with the average. Bars are color-coded: green for ≥0.9, yellow for 0.8–0.89, red for 0.7-0.79, and grey for < 0.7.}
\label{human_evaluation_result}
\end{figure}

\begin{table}[!t]
\centering
\caption{Qualitative Analysis of Question Categorization Methods. Here, the inputs question is \textquote*{\textbengali{টাইফয়েড কিভাবে ছড়িয়ে পড়ে?(How is typhoid spread?)}} and the actual category is  \textquote*{mode of transmission}.}
\label{tab:qualitative}
\begin{tabular}{|c|c|}
\hline
\textbf{Method} & \textbf{Predicted Category}\\
\hline
Perc & mode of transmission \\
\hline
MNB & mode of transmission\\
\hline
PAC & mode of transmission\\
\hline
RF & mode of transmission \\
\hline
SVM & mode of transmission \\
\hline
\end{tabular}
\label{ques_cat_ql_res}
\end{table}

\begin{table}[!t]
\centering
\caption{Demonstrating how anaphora resolution is performed in our system. Here for the second query \textquote*{\textbengali{এটির} (its)} refers to \textquote*{\textbengali{স্ট্রোক} (stroke)} according to the first query. Thus, \textquote*{\textbengali{এটির} (its)} is replaced by \textquote*{\textbengali{স্ট্রোক} (stroke)} to transform to modified second query.}
\begin{tabular}{|p{2.8cm}|p{6cm}|p{3.1cm}|}  
\hline
\textbf{Query} & \textbf{POS tagging} & \textbf{Modified} \\
\hline
\textbengali{স্ট্রোক কি?} (What is stroke?) & \textbengali{স্ট্রোক/NN-DIS কি/PRP ?/SYM)} (What/PRP is stroke/NN-DIS ?/SYM)) & \textbengali{স্ট্রোক কি?} (What is stroke?)\\
\hline
\textbengali{এটির লক্ষণ কি?} (What are its symptoms?) & \textbengali{এটির/PRP-OBJ লক্ষণ/NN-SYM কি/PRP ?/SYM)} What/PRP are its/PRP-OBJ symptoms/NN-SYM ?/SYM &  \textbf{\textbengali{স্ট্রোক}} \textbengali{লক্ষণ কি?} (What are the symptoms of \textbf{stroke}?)\\
\hline
\end{tabular}
\label{anaphora}
\end{table}

\begin{table}[!h]
\centering
\caption{Simulation of Question Answering System Interaction}
\label{qa_simulation}
\begin{tabular}{|p{6cm}|p{6.4cm}|}
\hline
\textbf{Simulation} & \textbf{English Translation} \\
\hline
\textbf{User:} \textbengali{ডায়েবেটিস  কি?} & \textbf{User:} What is diabetes? \\
\textbf{BanglaMed-QA:} \textbengali{ডায়েবেটিস একটি ক্রনিক মেডিক্যাল কন্ডিশন বা রোগ, যেখানে রক্তের গ্লুকোজ স্তর অস্বাভাবিক ভাবে বাড়তে থাকে।} & 
\textbf{BanglaMed-QA:} Diabetes is a chronic medical condition or disease in which blood glucose levels increase abnormally.\\
\hline
\textbf{User:} \textbengali{ডেঙ্গু কেন হয়}  & \textbf{User:} \textbengali{Why does dengue occur} \\
\textbf{BanglaMed-QA:} \textbengali{ডেঙ্গু এডিস মশার কামড়ে মানুষ ডেঙ্গু রোগে আক্রান্ত হয়ে থাকে। } & \textbf{BanglaMed-QA:} People are infected with dengue fever by the bite of the Aedes mosquito.\\
\hline
\textbf{User:} \textbengali{প্রতিরোধ??} & \textbf{User:} \textbengali{prevention??
} \\
\textbf{BanglaMed-QA:} \textbengali{ডেঙ্গু এর  প্রতিরোধ এ ঘরে যেন মশা ঢুকতে না পারে সে ব্যবস্থা করা৷ ঘরের আশপাশে বদ্ধ পানি জমা অথবা ময়লা ঝেটে দূর করুন। এমনকি ঘরে মশারি ব্যবহার করুন।} & 
\textbf{BanglaMed-QA:} Prevention of dengue: Make sure that mosquitoes cannot enter the house. Remove any stagnant water or dirt around the house by sweeping it away. Even use mosquito nets in the house.\\
\hline
\end{tabular}
\end{table}

The outcome of our model evaluation on the manually constructed test dataset is shown in Table \ref{tab:test_result}. The performance of base similarity methods is almost identical. But when we combine them, the performance is enhanced. The coverage in hard voting similarity decreases (6 were treated as NO\_PREDICTION), though base similarity methods show 100\% convergence. This is because of the use of an even number of individual base methods. Suppose, if two methods predicted $S_1$ and other two predicted $S_2$ as as the final query, then hard voting similarity can't be able to make a final decision and thus predicts that as NO\_PREDICTION. Fig. \ref{human_evaluation_result} shows the human evaluation result, ranging from 0.78 to 1.0. The average rating of 0.9 (out of 1.0) shows the system's overall consistently better, with most individuals rated highly. For this human evaluation, we have used soft voting similarity as it outperformed other methods (Table \ref{tab:test_result}).

\subsection{Qualitative analysis}
Table \ref{ques_cat_ql_res} shows that all classifier algorithms detect the category of questions correctly. Table \ref{anaphora} shows how the disambiguity of pronoun referencing is solved in our system. Here, the query with token having the POS tag PP-OBJ is replaced by the immediate antecedent noun. As seen from Table \ref{qa_simulation}, our \textquote*{BanglaMed-QA} Chatbot is taking queries about different diseases and providing brief information about their treatment, symptoms, prevention, and cures etc in conversational format that is very easy to understand.

\section{Conclusions}
\label{sec:conclusion}
The design of a question-answering system for medical texts in Bangla was carried out considering the challenges present in a low-resource language environment. We disambiguate pronoun references using POS tagging and rephrase questions with synonyms to help the system learn different configurations of interrogatory expressions. By mixing different methods of determining similarity and voting strategies, the system is intended to reduce the unreliability of providing answers. The categorization of questions is one technique that remarkably speeds up the entire process. We made provisions for evaluation, both automatic and manual, so that the quality of the answers can be thoroughly judged. This system opens up further developments and actual applications as a stepping stone for Bangla NLP in the medical field. In the future, we would like to integrate a disease prediction subsystem to make our system more robust. Moreover, we would like to explore more diseases and deep learning-based methods for query categorization and matching.

\bibliographystyle{splncs04}
\bibliography{mybibliography}

\appendix
\section{Interactive User Interface (UI)}
\begin{figure*}[htbp]
\centering
\centerline{\includegraphics[scale=0.33]{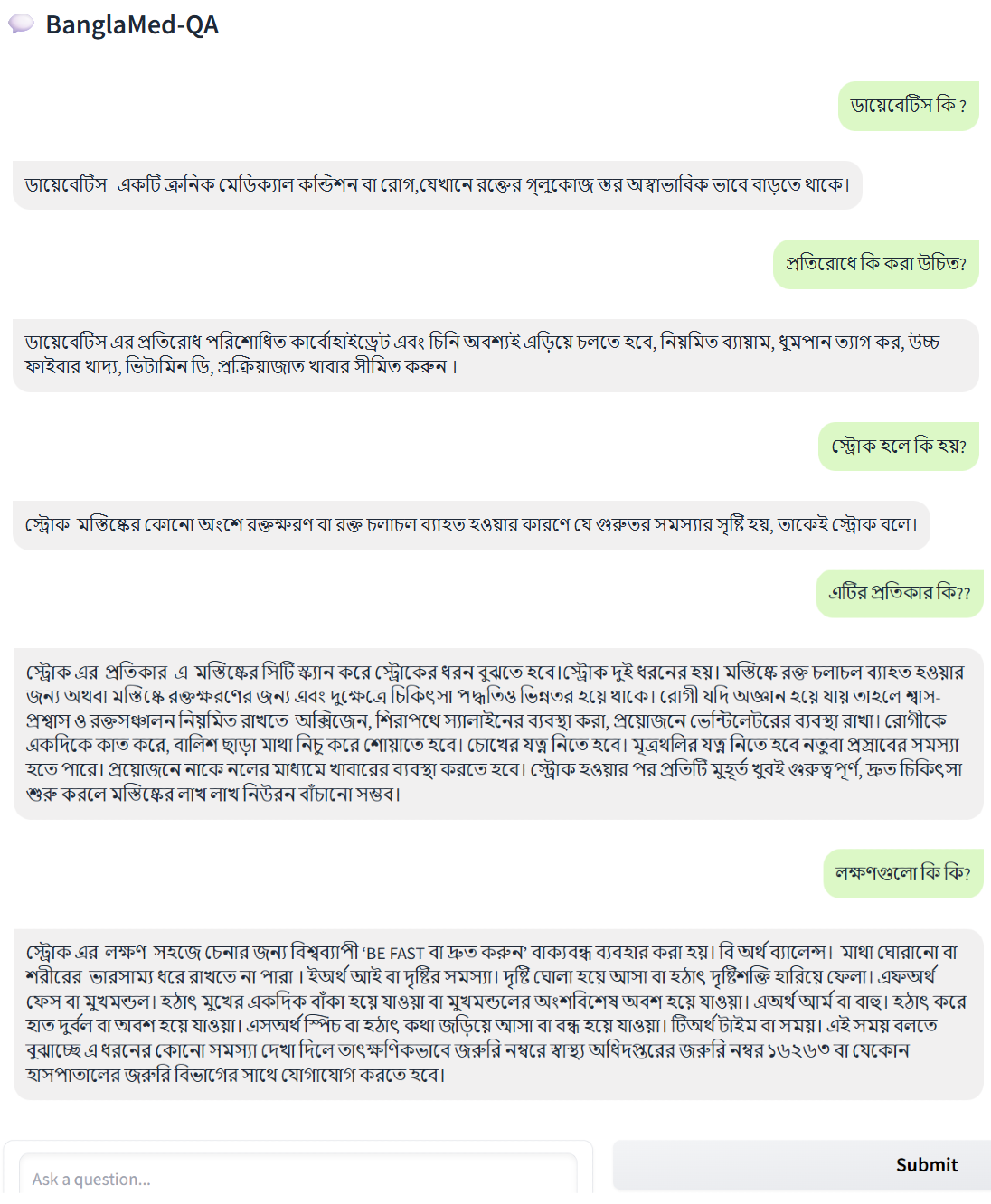}}
\caption{Interactive UI of our proposed BanglaMed-QA system.}
\label{simulation}
\end{figure*}

A practical system with an interactive UI to visualize the usefulness of our proposed system is shown in Fig. \ref{simulation}.

\end{document}